%% file: ex_article.tex
\documentclass[review,onefignum,onetabnum]{siamonline171218}

\usepackage{wrapfig}
\usepackage{amsmath}
\usepackage{color}
\usepackage{mdframed}
\newcommand{\method}{{\tt tHoops}}
\newcommand{\tensor}{\underline{\mathbf{X}}}

\newcommand{\silhouette}{\sigma}
\newcommand{\silhouettes}{\mathcal{S}}

\input{ex_shared}
\usepackage{multirow}
\usepackage{tcolorbox}
\usepackage{eurosym}
\tcbuselibrary{theorems}
\newtcbtheorem[number within=section]{mydefinition}{Model}%
{colback=yellow!5,colframe=yellow!35!black,fonttitle=\bfseries}{th}
\newtcbtheorem[number within=section]{mydefinition2}{Definition}%
{colback=blue!5,colframe=blue!35!black,fonttitle=\bfseries}{th}
\newtcbtheorem[number within=section]{mydefinition3}{Rule of Thumb}%
{colback=red!5,colframe=red!35!black,fonttitle=\bfseries}{th}
\usepackage{helvet}
\usepackage{courier}
\usepackage{epsf}
\usepackage{epsfig}
\usepackage{mathrsfs}
\usepackage{subfigure}
\usepackage{graphics}
\usepackage{booktabs}
\usepackage{balance}
\usepackage{graphicx}
\usepackage{balance}
\usepackage{epstopdf}
\usepackage{pifont}

\usepackage{amsfonts}

\usepackage{pifont}
\usepackage{xcolor}
\usepackage{epstopdf}
\usepackage{etoolbox}
\usepackage{tcolorbox}
\usepackage{tabularx}
\usepackage{array}
\usepackage{colortbl}
\tcbuselibrary{skins}

\DeclareRobustCommand{\officialeuro}{%
  \ifmmode\expandafter\text\fi
  {\fontencoding{U}\fontfamily{eurosym}\selectfont e}}

\newcolumntype{Y}{>{\raggedleft\arraybackslash}X}

\tcbset{tab1/.style={fonttitle=\bfseries\large,fontupper=\footnotesize\sffamily,
colback=yellow!10!white,colframe=red!75!black,colbacktitle=maroon!40!white,
coltitle=black,center title,freelance,frame code={
\foreach \n in {north east,north west,south east,south west}
{\path [fill=red!75!black] (interior.\n) circle (3mm); };},}}

\tcbset{tab2/.style={enhanced,fonttitle=\bfseries,fontupper=\footnotesize\sffamily,
colback=yellow!10!white,colframe=red!50!black,colbacktitle=maroon!40!white,
coltitle=black,center title}}

\tcbset{tab3/.style={enhanced,fonttitle=\bfseries,fontupper=\footnotesize\sffamily,
colback=yellow!10!white,colframe=red!50!black,colbacktitle=red!40!white,
coltitle=black,center title}}

\ifpdf
\hypersetup{
  pdftitle={Positional Value in Soccer: Expected League Points above Replacement},
  pdfauthor={Konstantinos Pelechrinis, Wayne Winston}
}
\fi

\myexternaldocument{ex_supplement}

\nolinenumbers
\begin{document}

\maketitle

\begin{abstract}
During the past few years advancements in sports information systems and technology has allowed the collection of a number of detailed spatio-temporal data that capture various aspects of basketball. 
For example, shot charts, that is, maps capturing locations of (made or missed) shots, 
and spatio-temporal trajectories for the players on the court can capture information about the offensive and defensive tendencies, as well as, schemes used by a team. 
Characterization of these processes is important for player and team comparisons, scouting, game preparation etc. 
Team and player tendencies have traditionally been compared in a heuristic manner, which inevitably can lead to subtle but crucial information being ignored. 
Recently automated ways for these comparisons have appeared in the sports analytics literature. 
However, these approaches are almost exclusively focused on the spatial distribution of the underlying actions (usually shots taken), ignoring a multitude of other parameters that can affect the action studied.  
In this study, we propose a framework based on tensor decomposition for obtaining a set of prototype spatio-temporal patterns based on the core spatio-temporal information and contextual meta-data. 
At the epicenter of our work is a 3D tensor $\tensor$, whose dimensions represent the entity under consideration (team, player, possession etc.), the location on the court and time. 
We make use of the PARAFAC decomposition and we decompose the tensor into several interpretable patterns, that can be thought of as prototype patterns of the process examined (e.g., shot selection, offensive schemes etc.).  
We also introduce an approach for choosing the number of components to be considered. 
Using the tensor components, we can then express every entity as a weighted combination of these components. 
Finally, the framework introduced in this paper has applications that go beyond purely pattern analysis. 
In particular, it can facilitate a variety of tasks in the work-flow of a franchise's basketball operations as well as in the sports analytics research community. 
\end{abstract}

\section{Introduction}
\label{sec:intro}

What are the offensive tendencies of your upcoming opponent with regards to their shot selection? 
Do these tendencies change through the course of the game? 
Are they particularly ineffective with regards to specific plays so as to force them towards them? 
These are just some of the questions that our proposed framework, named {\method}, can answer. 
While data have been an integral part of sports since the first boxscore was recorded in a baseball game during the 1870s, it is only recently that machine learning has really penetrated the sports industry and has been utilized for facilitating the operations of sports franchises.  
One of the main reason for this is our current ability to collect more fine-grained data; data that capture essentially (almost) everything that happens on the court.

\begin{figure*}[ht]
\centering
\includegraphics[scale=0.3]{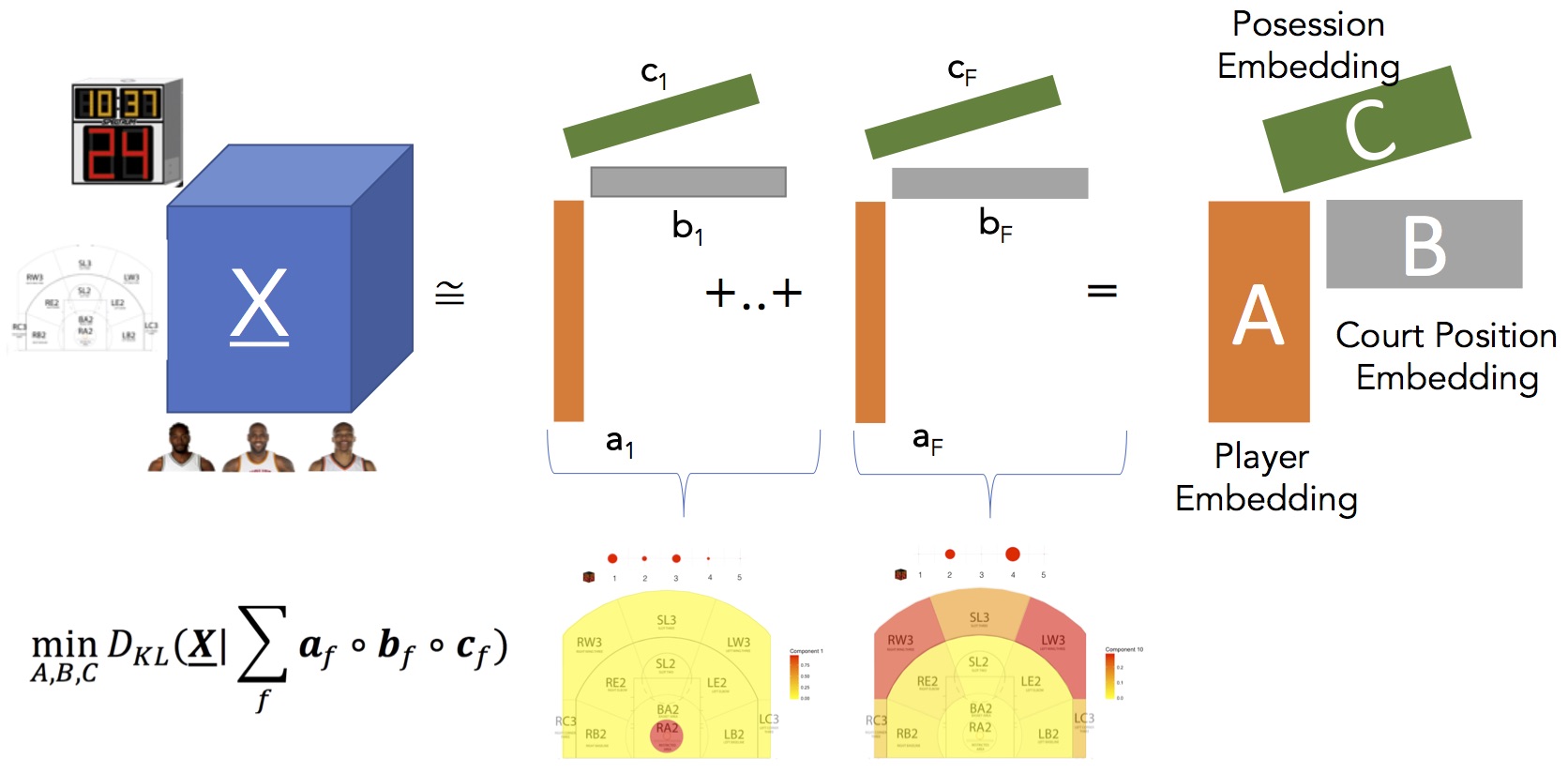}
\caption{The {\method} framework. Tensor $\tensor$ can capture the aggregate shot charts of each players for specific times within the game. For example, $\tensor(i,j,k)$ is the number of shots that player $i$ took from court location $j$ during time $k$. $\method$ identifies prototype patterns in the data, expressed as triplets of vectors corresponding to the three dimensions of $\tensor$ respectively. An element in the player vector can be thought of as a {\em soft} coefficient for the membership of the corresponding player in this component/pattern.}
\label{fig:thoops}
\end{figure*}

For example, shot charts, that is, maps capturing locations of (made or missed) shots, describe the shot selection process and can be thought of as an indicator of the identity of a player/team. 
Furthermore, since the 2013-14 season, the National Basketball Association (NBA) has mandated its 30 teams to install an optical tracking system that collects information 25 times every second for the location of all the players on the court, as well as the location of the ball. 
These data are further annotated with other information such as the current score, the game and shot clock time etc. 
Optical tracking data provide a lens to the game that is much different from traditional player and team statistics. 
These spatio-temporal trajectories for all the players on the court can capture information for the offensive/defensive tendencies as well as, the schemes used by a team. 
They can also allow us to quantify parts of the game that existing popular statistics cannot. 
For instance, the Toronto Raptors were among the first teams to make use of this technology and were able to identify optimal positions for the defenders, given the offensive formation \cite{grandland}.  
This allowed the Raptors to evaluate the defensive performance of their players, an aspect of the game that has traditionally been hard to evaluate through simple boxscore metrics such as blocks and steals. 

One of the tasks a team has to undertake during its preparation for upcoming games is to study its opponents, their tendencies and how they compare with other teams. 
Playing tendencies have been traditionally analyzed in a heuristic manner, mainly through film study, which is certainly time consuming (a temporal cost pronounced for NBA teams that play 3 to 4 games every week). 
However, the availability of detailed spatio-temporal (optical tracking) data makes it possible to identify prototype patterns of schemes and tendencies of the opponent in much shorter time. 
Recently automated ways for similar comparisons have appeared in the sports analytics literature focusing on shooting behavior and aiming into identifying a set of prototype shooting patterns that can be used as a basis for describing the shooting tendencies of a player/team (e.g., \cite{miller14}). 
These approaches offer a number of advantages over simply comparing/analyzing the raw data (e.g., the raw shot-charts). 
In particular, similar to any latent space learning method, they allow for a better comparison as well as easier data retrieval, through the decomposition of the data into several prototype patterns in a reduced dimensionality. 
However, existing approaches almost exclusively focus on the spatial distribution of the underlying (shooting) process ignoring a multitude of other parameters that can affect the shot selection. 
For instance, the time remaining on the shot/game clock, the score differential, etc. are some contextual factors that can impact the shot selection of a player.  
Similarly, the analysis of the players' trajectories obtained from optical tracking can benefit greatly from the learning of a latent, reduced dimensionality, space that considers several contextual aspects of the game.

\begin{mdframed}[linecolor=red!60!black,backgroundcolor=gray!20,linewidth=2pt,  topline=true,  rightline=false, leftline=false] 
To address this current gap in the literature, we design and develop {\method}, a novel tensor decomposition based method for analyzing basketball data, which simultaneously incorporates multiple factors  that can affect the offensive (and defensive) tendencies of a team. 
\end{mdframed}

{\method} allows us to separate the observed tendencies captured in the data across multiple dimensions and identify patterns strongly connected among all dimensions. 
This is particularly important for sports, where strategies, play selection and schemes depend on several contextual factors including time. 
What are the Boston Celtics offensive patterns in their ATO (After Time-Out) plays? 
How do they differ based on the players on the court? 
The benefits of {\method} are not limited to describing and learning the tendencies of a team/player but as we will discuss later it also allows for flexible retrieval of relevant parts of the data from the coaching staff (e.g., for film study). 
We would like to emphasize here that we do not suggest that {\method} - or any other similar system - will substitute film study, but it will rather make it more efficient by allowing coaching staff to (i) obtain a report with the most prevalent patterns, which can form a starting point for the film study, and (ii) perform a flexible search through the data (e.g., identify all the possessions that had two offensive players at the corner threes and one player in the midrange slot during their last 5 seconds). 
Simply put, {\method} is an automated, {\em exploratory analysis} and {\em indexing}, method that can facilitate traditional basketball operations.  
In addition, {\method} can be used to generate synthetic data based on patterns present in real data. 
This can significantly benefit the sports analytics community, since real optical tracking data are kept proprietary. 
However, as we will describe later, one could identify prototype motifs on the real data using {\method} and use the obtained patterns to generate synthetic data that exhibit similar patterns. 

In this study we focus on and evaluate the analytical tool of {\method} but we also describe how we can use {\method} for other applications (i.e., indexing player tracking data to facilitate flexible retrieval and generating synthetic data).   
More specifically, we are interested in analyzing the offensive tendencies of an NBA team (or the league as a whole). 
For capturing these offensive tendencies we make use of two separate datasets, namely, {\em shot charts} and {\em optical tracking} data. 
The reason for this is twofold. 
First, the two different datasets capture different type of information related with the offensive tendencies of a team.  
The shot charts are representative of the shot selection process for a team or a player, while the optical tracking data capture rich information for the shot generation process, i.e., the players' and ball movement that led to a shot (or a turnover). 
Second, using two datasets that encode different type of information directly showcases the general applicability of {\method}, i.e., it can be adjusted accordingly to accommodate a variety of multi-aspect (sports) data.

In brief, {\method} is based on tensor decomposition (see Figure \ref{fig:thoops}).  
For illustration purposes, let us consider the shot charts of individual players. 
{\method} first builds a 3-dimensional tensor $\tensor$, whose element $\tensor(i,j,k)$ is the aggregate number of shots that player $i$ took from court location $j$ at time $k$. 
The granularity of location and time can be chosen based on the application. 
For example, for location one could consider a grid over the court and hence, $j$ represents a specific grid cell.
Or alternatively - as is the case in Figure \ref{fig:thoops} - $j$ can represent one of the official courtzones (e.g., left/right corner three area, left/right slot three area etc.).  
With respect to the temporal dimension, one could consider as the time unit to be the shot clock (i.e., values between 0-24), the quarter in the game (i.e., values between 1-5, where 5 groups together all possible overtimes) or even the exact game clock (e.g., at the minute granularity).  
The factors of this tensor, which are obtained through solving an optimization problem, are essentially vector triplets, that can also be represented as 3 separate matrices (Figure \ref{fig:thoops}). 
As we will elaborate on later, these factors provide us with a set of prototype patterns, i.e., {\em shooting bases}, that synthesize the shot selection tendencies of players. 
From a technical standpoint, one of the challenges is to identify the appropriate number of factors/components for the decomposition. 
While, there are well-established metrics for this task, they have limitations - both in terms of computational complexity as well as in terms of applicability - that can appear in our setting. 
Therefore, we introduce an approach that is based on a specifically-defined clustering task and the separability of the obtained clusters. 


The rest of the paper is organized as follows: 
Section \ref{sec:related} provides an overview of related with our work literature. 
We further present {\method} in Section \ref{sec:thoops}. 
Sections \ref{sec:dataset} and \ref{sec:thoops-results} describe our datasets and their analysis using {\method} respectively. 
Finally, Section \ref{sec:discussion} discusses our work and other potential applications of {\method}. 

\section{Related Literature}
\label{sec:related}

The availability of optical tracking sports data has allowed researchers and practitioners in sports analytics to analyze and model aspects of the game that were not possible with traditional data. 
For example, Franks \textit{et al.} \cite{franks15} developed  models for capturing the defensive ability of players based on the spatial information obtained from optical tracking data. 
Their approach is based on a combination of spatial point processes, matrix factorization and hierarchical regression models and can reveal several information that cannot be inferred with boxscore data. 
For instance, the proposed model can identify whether a defender is effective because he reduces the quality of a shot or because he reduces the frequency of the shots all together. 
Cervone \textit{et al.} \cite{cervone2016} further utilize optical tracking data and develop a model for the expected possession value (EPV) using a multi-resolution stochastic process model. 
Tracking the changes in the EPV as the possession progresses can enable practitioners to quantify previously \textit{intangible} contributions such as a good screen, a good pass (not assist) etc. 
Similar to this study, Yue \textit{et al.} \cite{yue2014learning} develop a model using conditional random fields and non-negative matrix factorization for predicting the near-term actions of an offense (e.g., pass, shoot, dribble etc.) given its current state.  
In a tangential direction, D'Amour \textit{et al.} \cite{damour15} develop a continuous time Markov-chain to describe the discrete states a basketball possession goes through. 
Using this model the authors then propose entropy-based metrics over this Markov-chain to quantify the ball movement through the unpredictability of the offense, which also correlates well with the generation of opportunities for open shots.  
Optical tracking data can also quantify the usage of the different court areas. 
Towards this direction Cervone \textit{et al.} \cite{cervone2016nba} divided the court based on the Voronoi diagram of the players' locations and formalized an optimization problem that allowed them to obtain court realty values for different areas of the court. 
This further allowed the authors to develop new metrics for quantifying the spacing of a team and the positioning of the lineup. 
Very recently a volume of research has appeared that utilizes deep learning methods to analyze spatio-temporal basketball data and learn latent representations for players and/or teams, identify and predict activities etc. (e.g., \cite{Mehrasa18,zhong18} with the list not being exhaustive).

Closer to our work, Miller \textit{et al.} \cite{miller14} use Non-Negative Matrix Factorization to reduce the dimensionality of the spatial profiles of player's shot charts. 
Their main contribution is the use of a log-Gaussian Cox point process to smooth the raw shooting charts, which they show provides more intuitive and interpretable patterns. 
The same authors developed a dictionary for trajectories that appear in basketball possessions using Bezier curves and Latent Dirichlet Allocation \cite{miller2017possession}. 
Our work can be thought of as complementary to these studies. 
In particular, {\method} is able to consider additional dimensions that can affect the shot selection process, and the possession development such as the time while it is also able to analyze a large variety of multi-aspect in general sports data and is not limited on shooting charts and spatial trajectories. 
Furthermore, is generic enough to power other applications (see Section \ref{sec:discussion}).

While basketball is the sport that has been studied the most through optical tracking data - mainly due to the availability of data - there is relevant literature studying other sports as well (both in terms of methodology and application). 
For example, Bialkowski \textit{et al.} \cite{bialkowski2014large} formulate an entropy minimization problem for identifying players' roles in soccer.  
They propose an EM-based scalable solution, which is able to identify the players' role as well as the formation of the team. 
Lucey \textit{et al.} \cite{lucey2014quality} also used optical tracking data for predicting the probability of scoring a goal by extracting features that go beyond just the location and angle of the shot \cite{fairchild17}. 
More recently, Le \textit{et al.} \cite{Le2017CoordinatedMI} develop a collaboration model for multi-agents using a combination of unsupervised and imitation learning. 
They further apply their model to optical tracking data from soccer to identify the optimal positioning for defenders - i.e., the one that \textit{minimizes} the probability of the offense scoring a goal - given a particular formation of the offense.  
This allows teams to evaluate the defensive skills of individual players. 
In a tangential effort, Power \textit{et al.} \cite{Power:2017:PCE:3097983.3098051} define and use a supervised learning approach for the risk and reward for a specific pass in soccer using detailed spatio-temporal data. 
The risk/reward for a specific pass can further quantify offensive and defensive skills of players/teams. 
While we introduce {\method} as a framework for analyzing basketball data, it should be evident that it can really be used to analyze spatio-temporal (and in general multi-aspect) data for other sports as well.

\section{\method: Tensor Representation and Decomposition}
\label{sec:thoops}

In this section we will present the general representation of spatio-temporal sports data with tensors, as well as, the core of {\method}. 
A $n$-mode tensor, is a generalization of a matrix (2-mode tensor) in $n$ dimensions. 
For illustration purposes in this section we will focus on players' shot charts that include information for the court location, game time and the player who took the shot. 
In order to represent these shot charts, we will utilize a 3-mode tensor $\tensor$, that captures the spatio-temporal information of the shot selection for players/teams. 
In particular, the element $\tensor(i, j, k)$ will be equal to the number of shots that player/team $i$ took from the court location $j$ during time $k$. 
Figure \ref{fig:thoops} depicts this (cubic) structure. 
A typical technique for identifying latent patterns in data represented by a 2-mode tensor (i.e., a matrix), is matrix factorization (e.g., Singular Value Decomposition, Non-negative Matrix Factorization etc.). 
A generalization of the Singular Value Decomposition in n-modes is the Canonical Polyadic (CP) or PARAFAC decomposition \cite{harshman1970foundations}. 
Without getting into the details of the decomposition, PARAFAC expresses $\tensor$ as a sum of $F$ rank-one components: 

\begin{equation}
\tensor \approx \displaystyle {\sum_{f=1}^F  \mathbf{a}_f \circ \mathbf{b}_f \circ \mathbf{c}_f },
\label{eq:tesor_dec}
\end{equation}
where $ \mathbf{a}_f \circ \mathbf{b}_f \circ \mathbf{c}_f (i,j,k) = \mathbf{a}_f(i) \mathbf{b}_f (j) \mathbf{c}_f (k) $. 
In cases where we have sparse count data these components are obtained as the solution to the following optimization problem \cite{chi2012tensors}:

\begin{equation}
\min_{{\mathbf{A}},\mathbf{B},\mathbf{C}} D_{KL}(\tensor|{\sum_{f}  \mathbf{a}_f \circ \mathbf{b}_f \circ \mathbf{c}_f }),
\end{equation}
where $D_{KL}$ is the Kullback-Leibler divergence, and matrices $\mathbf{A,B,C}$ hold the $\mathbf{a}_f,\mathbf{b}_f,\mathbf{c}_f$ respectively in their $f$-th columns. 
Simply put, each component of the decomposition, i.e., triplet of vectors, is a rank-one tensor (obtained as the outer product of the three vectors). 
Each vector in the triplets corresponds to one of the three modes of the original tensor $\tensor$. 
In our example, $\mathbf{a}$ corresponds to the players, $\mathbf{b}$ corresponds to the court locations, and $\mathbf{c}$ corresponds to the game clock/time.
Each of these $F$ components can be considered as a cluster, and the corresponding vector elements as soft clustering coefficients, that is, if a coefficient is small, the corresponding element does not belong to this {\em cluster}. 
In our application, these \textit{clusters} correspond to a set of players that tend to take shots from \textit{similar} areas on the court during \textit{similar} times within the game. 
For notational simplicity, we will denote as matrix $\mathbf{A}$ (and matrices $\mathbf{B}$ and $\mathbf{C}$ accordingly) the factor matrix that contains the $\mathbf{a}_f$ ($\mathbf{b}_f$ and $\mathbf{c}_f$ respectively) vectors as columns.
The vectors ($\mathbf{b}$, $\mathbf{c}$) essentially correspond to the latent patterns for the spatio-temporal shot selection of players obtained from tensor $\tensor$. 

\textbf{Intuition Behind the Use of Tensors: }
Tensor decomposition attempts to summarize the given data into a reduced rank representation. 
PARAFAC tends to favor dense groups that associate all the aspects involved in the data (player, locations and time in the example in Figure \ref{fig:thoops}). 
These groups need not be immediately visible via inspection of the $n$-mode tensor, since PARAFAC is not affected by permutations of the mode indices. 
As an immediate consequence, we expect near-bipartite cores of players who take shots from specific locations on the court during certain periods of the game. 
The benefit of tensor decomposition over matrix decomposition - that has been used until now to analyzing shooting patterns - is the ability to consider several aspects of the data simultaneously, allowing for a richer context, consequently allowing {\method} to obtain a richer set of latent patterns. 
One could argue that we could compare shot-charts directly after dividing them based on the game time of the shot (or other contextual factor).  
This is indeed true, but as aforementioned the dimensionality of the raw data can be very high (especially with an increase in the contextual factors considered), which will make it challenging to identify high quality patterns. 

\textbf{Choice of Number of Components $F$: }
Depending on the structure of the given data, the PARAFAC decomposition can range from (almost) perfectly capturing the data, to performing rather poorly. 
The main question is whether the spatio-temporal data at hand are amenable to PARAFAC analysis, and to what extent. 
In order to answer the question of how well does PARAFAC decomposition model our data, we turn our attention to a very elegant diagnostic tool, CORCONDIA \cite{bro03}. 
CORCONDIA serves as an indicator of whether the PARAFAC model describes the data well, or whether there is some problem with the model. 
The diagnostic provides a number between 0 and 100; the closer to 100 the number is, the better the modeling. 
If the diagnostic gives a low score, this could be caused either because the chosen rank $F$ is not appropriate, or because the data do not have appropriate trilinear structure, regardless of the rank. 
To identify the reason behind a low CORCONDIA score one can gently increase the rank and observe the behavior \cite{papalexakis2015location}.

Despite its elegant application, computing CORCONDIA is very challenging even for moderately large size data. 
The main computational bottleneck of CORCONDIA is solving the following linear system:
$
	\mathbf{g} = \left(  \mathbf{C}\otimes \mathbf{B} \otimes \mathbf{A} \right)^\dag vec\left(\tensor\right)
$
where $\dag$ is the Moore-Penrose pseudoinverse, $\otimes$ is the Kronecker product, and the size of $\left(  \mathbf{C}\otimes \mathbf{B} \otimes \mathbf{A} \right)$ is $IJK \times F^3$. 
Even computing and storing $\left(  \mathbf{C}\otimes \mathbf{B} \otimes \mathbf{A} \right)$ proves very hard when the dimensions of the tensor modes are growing, let alone pseudoinverting that matrix. 

In order to tackle the above inefficiency in this study we use our recent work for efficiently computing CORCONDIA when the data are large but sparse \cite{papalexakis2015fastcorcondia}. 
In brief, key behind the approach is avoiding to pseudoinvert $\left( \mathbf{A \otimes B \otimes C} \right)$.  
In order to achieve the above, we reformulate the computation of CORCONDIA.
The pseudoinverse 
$
\left( \mathbf{A} \otimes \mathbf{B} \otimes \mathbf{C} \right)^\dag
$
can be rewritten as: 

\begin{equation}
\left( \mathbf{V_a} \otimes \mathbf{V_b} \otimes \mathbf{V_c} \right) \left( \mathbf{\Sigma_a}^{-1} \otimes \mathbf{\Sigma_b}^{-1} \otimes \mathbf{\Sigma_c}^{-1} \right)  \left( \mathbf{U_a}^T \otimes \mathbf{U_b}^T \otimes \mathbf{U_c}^T \right) 
\end{equation}
where $\mathbf{A} = \mathbf{U_a \Sigma_a {V_a}^T}$, $\mathbf{B} = \mathbf{U_b \Sigma_b {V_b}^T}$, and $\mathbf{C} = \mathbf{U_c \Sigma_c {V_c}^T}$ (i.e. the respective Singular Value Decompositions).
After rewriting the least squares problem as above, we can efficiently compute a series of Kronecker products times a vector, {\em without} the need to materialize the (potentially big) Kronecker product.

One of the limitations of CORCONDIA is that it cannot examine the quality of the decomposition for rank higher than the smallest dimension of the original tensor $\tensor$.  
Depending on the specific design of the tensor for {\method}, the smallest dimension of $\tensor$ can limit the practical  applicability of CORCONDIA for choosing rank $F$. 
Simply put, with $f$ being the smallest dimension of $\tensor$, using CORCONDIA will provide us with a rank $F$ of at most $f$, i.e., $F \le f$. 
If the CORCONDIA score has already been reduced for a decomposition of rank $r \le f$ then we do not have to further examine other ranks, since their quality is going to be low. 
However, if the CORCONDIA score is still high for a decomposition of rank $f$, a higher rank decomposition can provide us with a more practical answer that captures a larger percentage of the patterns that exist in the data. 
In order to overcome this problem in these cases we introduce a heuristic approach that attempts to choose the rank $F$ of the decomposition based on the ability of the identified components to separate natural clusters in the data as compared to the raw data. 

More specifically let us consider tensor $\tensor$ whose first mode (say mode $A$) is the dimension across which we want to separate the data (e.g., players, teams etc.).  
In other words, the second and third mode ($B$ and $C$ respectively), will be the feature space that we will use for the clustering. 
First we perform clustering across the first mode using the raw data, i.e., the features for element $i$ of the first mode is the matrix $\tensor(i,:,:)$. 
In order to quantify the quality of the clustering we can use the Silhouette \cite{rousseeuw1987silhouettes} of the obtained clustering $\silhouette_{\tensor,k}$, where $k$ represents the number of clusters. 
Given that we do not know the number of clusters a-priori we compute a set of Silhouette values for different number of clusters  $\silhouettes_{\tensor,2:K}$. 
From these values, one can choose their average, their minimum or their maximum value. 
In our work, we choose to pick as the Silhouette value the $\max \silhouettes_{\tensor,2:K}$. 
The reasoning behind this choice, is that the maximum Silhouette value will provide the best separation of the data. 
Consequently for the rank $F$ decomposition of $\tensor$ we  consider as the features $\mathbf{r}$ for the element $i$ of the first mode, a concatenation of the corresponding elements in the component vectors $\mathbf{a}$ (or simply the $i^{th}$ row of the factor matrix $\mathbf{A}$):

\begin{equation}
\mathbf{r}_i = (\mathbf{a}_{i,j},~\forall j\in\{1,\dots,F\})
\label{eq:dec_f}
\end{equation}
Using these features we can again cluster the elements in the mode of interest of the tensor and obtain a set of Silhouette values $\silhouettes_{F,2:K}$.  
We can then choose the decomposition rank $F$ as: 

\begin{equation}
\min_F \{\max{\silhouettes}_{\tensor,2:K} \le \max{\silhouettes}_{F,2:K} \land |\max{\silhouettes}_{F+1,2:K} - \max{\silhouettes}_{F,2:K}| < \epsilon \}
\label{eq:heuristic}
\end{equation}
where $\epsilon > 0$ is a convergence criteria. 
Simply put we choose the decomposition rank $F$ that provides better separability as compared to the raw data, while at the same time increasing the rank does not provide any (significant) additional benefits. 
We can also add another constraint in Equation \ref{eq:heuristic} that sets a minimum value for  $\max{\silhouettes}_{F,2:K}$.  
According to Kaufman and Rousseeuw \cite{kaufman2009finding} a strong structure will exhibit Silhouette values greater than 0.7.  
Therefore, we could set a threshold according to similar rules-of-thumb. 
Nevertheless, this could be very restrictive in some cases (e.g., when there is not any inherent natural structure in the data), and therefore, we do not include it in the formal condition in Equation \ref{eq:heuristic}. 

\section{Datasets}
\label{sec:dataset}

In this section we present the datasets we used in our study as well as the results from our analysis. For tensor manipulations we use the Tensor Toolbox for Matlab \cite{TTB_Software}.

\begin{figure}[ht]
\centering
\includegraphics[scale=0.4]{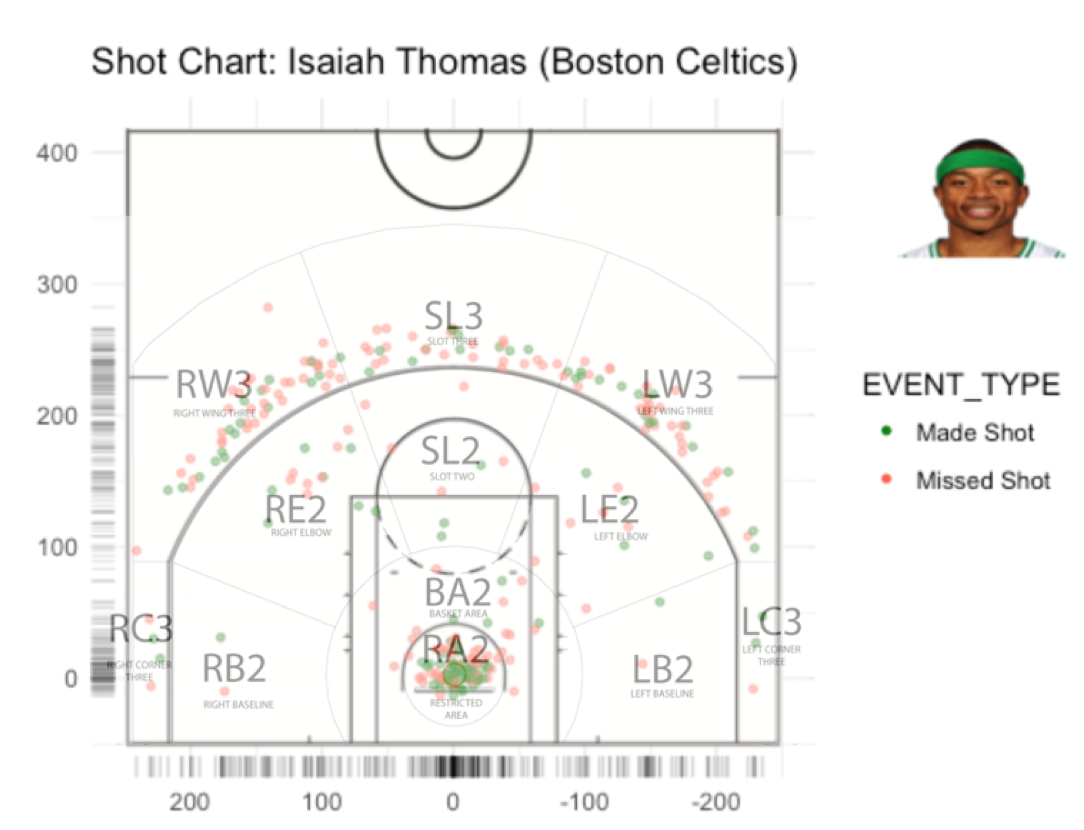}
\caption{Sample shotchart, with the court zones used for {\method} spatial granularity depicted.}
\label{fig:shotchart}
\end{figure}

{\bf Shot Charts: }
We collected a shot dataset from the 2014-15 NBA season using NBA's shotchart API endpoint \cite{kpele-git-2018}. 
This endpoint provides several information for all the shots taken during the season, including: the game that the shot was taken, the player that took the shot, the location on the floor from where the shot was taken, the game clock information, the shot type, and whether the shot was made or missed. 
In total we collected information for 184,209 shots from 348 different players.
In order to represent these data with a tensor $\tensor$, we need to quantize the court location and the temporal dimension. 
For court location we could filter the points through a spatial grid and use the grid ID as the index for the location dimension in $\tensor$. 
However, we choose to use the official courtzones,  depicted in the sample shotchart in Figure \ref{fig:shotchart}, 
since these are the natural borders for the different locations on the court. 
This choice further reduces any potential noise induced by an extremely fine-grained grid.  
Furthermore, for the temporal dimension, we use the game periods, where we merge all overtimes to a single 5th period.

{\bf Player Tracking: }
Since 2013 the NBA has \textit{mandated} all the teams to equip their stadiums with optical tracking capabilities. 
The information this technology provides is derived from cameras mounted in stadium rafters and consist primarily of x,y coordinates for the players on the court and ball. 
This information is recorded at a frequency of 25 times per second and allows the analysis of offensive (and defensive) schemes. 
The player tracking data provide additional meta-information such as game and shot clock, violations etc. 
For our study, we use data from 612 games from the 2015-16 NBA regular season \cite{linouk23-git-2016}. 
As we will see in the following section, these data can be analyzed by {\method} to provide prototype patterns for the offensive tendencies of a team, which can further facilitate scoutings tasks as discussed earlier. 

\section{Evaluations}
\label{sec:thoops-results}

In this section we present the results from the application of {\method} to our datasets. 
We would like to emphasize here that {\method} is an unsupervised learning method and given the absence of ground truth labels (i.e., the {\em true patterns}) it is hard to do a comparative/accuracy evaluation.	 
Therefore, similar to existing literature that deals with related unsupervised learning problems \cite{miller14,miller2017possession}, we perform more of a qualitative evaluation of the results, discussing and matching them with {\em partial} ground truth that we know for players and teams. 

\begin{figure}[ht]
\centering
\includegraphics[scale=0.3]{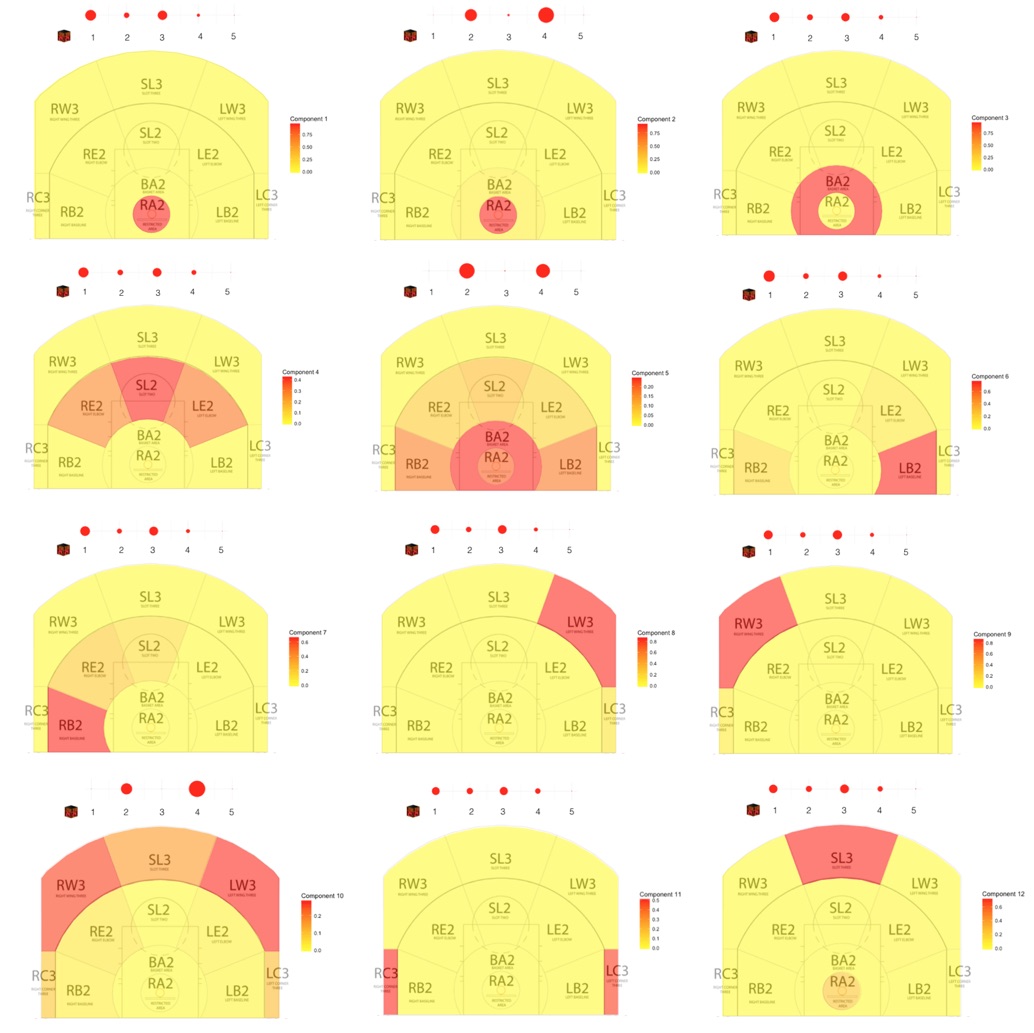}
\caption{{\method} components for the $\tensor_{Made}$ tensor. The spatial and temporal elements are presented. The size of the points for the temporal element correspond to the coefficient for each quarter.}
\label{fig:shot-components}
\end{figure}

\textbf{Shot Chart Analysis: }
We start by presenting our results for the players' latent shooting patterns. 
In particular, we build two separate tensors, one for made shots $\tensor_{Made}$ and one for missed shots $\tensor_{Missed}$, since one can argue that they encode different information of sorts. 
Figure \ref{fig:shot-components} presents the spatial and temporal patterns for the 12 components we identified using the $\tensor_{Made}$. 
Note here that the smallest dimension of $\tensor_{Made}$ is the temporal one and this is equal to 5. 
Therefore, CORCONDIA can assess the quality of the decomposition for rank up to 5.  
Our results indicate that the quality of the model obtained does not deteriorate until that rank so we also examine higher ranks. 
However, PARAFAC provides us only with 12 components for $\tensor_{Made}$, that is, higher components are degenerate, and for this reason we use all of the components provided. 

Components 1 and 2 are particularly important for showing the difference between {\method} and a similar approach based on matrix factorization. 
The spatial element of these two components is very similar (almost identical) and represents shots made from the (deep) paint. 
However, they are different in the temporal dimension. 
As we can see component 1 includes shots taken mainly during quarters 1 and 3, while component 2 mainly covers quarters 2 and 4. 
The fact that PARAFAC detected two components (instead of of one that covers all the periods), means that there are subgroups of players that take and make these shots in different times during the game. 
Of course, this difference can be purely based on personnel decisions from the coaching staff through the game, but {\method} is able to pick this up and provide us with latent patterns considering all the dimensions included in the tensor simultaneously. 
In contrast, component 11 corresponds to corner 3 (made) shots. 
There is no other component that includes them (component 10 includes a small fraction of corner 3 shots but it heavily captures above the break 3-point shots), which means that players that are heavily represented in this component take (and make) these shots almost uniformly across the game as it can be seen by the temporal element.

Figure \ref{fig:shot-components} does not provide any information for the player vector of the component. 
The player vector of the tensor factor informs us which players have a strong representation in the component under examination. 
For example, Table \ref{tab:players} presents the top-10 players (i.e., the 10 players with the largest coefficients in the corresponding player vectors) included in the corner 3 components and the midrange shot component.

\begin{table}[h!]
  \begin{center}
    \begin{tabular}{c|c} 
      \textbf{Corner 3} & \textbf{Midrange} \\
       Component 11 & Component 4\\ 
      \hline
      Trevor Ariza & Blake Griffin\\
      Matt Barnes & Avery Bradley \\
      Danny Green & Monta Ellis \\
      Klay Thomson & LaMarcus Aldridge \\
      Luol Deng & Anthony David \\
      Kyle Korver & Marc Gasol \\
      JJ Redick & Pau Gasol \\
      O.J. Mayo & Nikola Vucevic\\
      Bojan Bojdanovic & Chris Paul\\
    \end{tabular}
  \end{center}
      \caption{The components obtained from {\method} can provide us with valuable information for the shooting tendencies of players.}
          \label{tab:players}
          \vspace{-0.1in}
\end{table}

\begin{figure}[ht]
\centering
\includegraphics[scale=0.3]{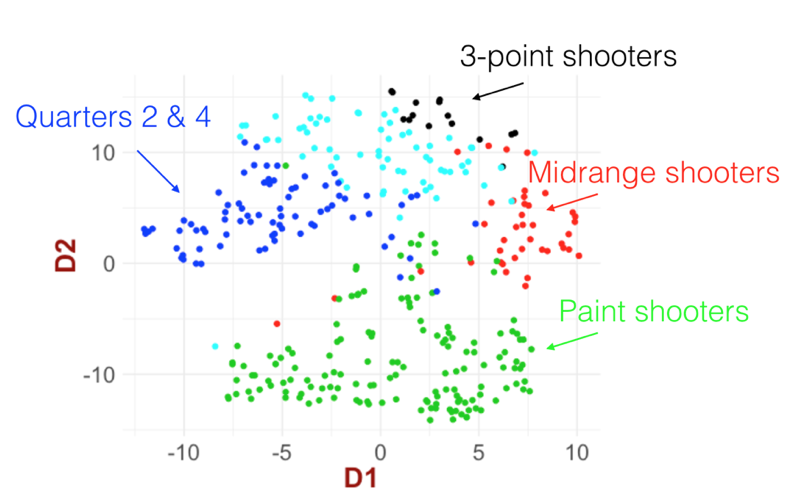}
\caption{t-SNE visualization of the players' clusters using the components for $\tensor_{Made}$ obtained through {\method}.}
\vspace{-0.1in}
\label{fig:player_clusters}
\end{figure}

As one might have expected, Danny Green, Klay Thompson, Kyle Korver and JJ Redick are predominantly featured in the corner 3s component, while players like LaMarcus Aldridge, Chris Paul and the Gasol brothers are featured in the midrange component. 
Table \ref{tab:players} also serves as an indicator that the components obtained from {\method} are sensible and pass the ``eye-test''. 
Using as the players features the vectors $\mathbf{r}$ from Equation (\ref{eq:dec_f}), we can obtain a 12-dimensional latent representation of each player that can be further used to cluster players. 
These clusters will represent players with similar offensive patterns (with regards to shots made). 
We use k-means clustering and the Silhouette to determine the appropriate number of clusters, which provides us with a value of k=5. 
Figure \ref{fig:player_clusters} further presents the clusters on a two-dimensional projection using t-SNE \cite{maaten2008visualizing}. 
As we can see the clusters are well distinguished - especially considering that t-SNE uses a (further) reduced dimensionality of the data. 
The largest cluster corresponds to players whose main patterns (i.e., the ones with the highest coefficients) correspond to shots taken from the paint (specifically tensor components 1, 2, and 3). 
The smallest cluster corresponds to players whose patterns heavily include the 3-point shoot components (tensor components 8, 9, 11 and 12). 
This cluster includes players such as Steph Curry, James Harden, Kyle Korver, JJ Redick, Gordon Hayward, Kyrie Irving, Klay Thompson and JR Smith. 
Another distinct cluster includes players whose most dominant components are 4, 6 and 7, i.e., midrange shots. 
This cluster includes players like DeMar DeRozan, LaMarcus Aldridge, Al Horford, Blake Griffin, Marreese Speights and Anthony Davis. 
The fourth cluster does not exhibit any specific pattern with regards to the spatial distribution of the shots.
However, it includes players who are offensively active mainly during quarters 2 and 4 (tensor components 2, 5 and 10). 
Players that fall into this cluster are mainly bench and role players such as Jamal Crawford, Leandro Barbosa, Patty Mills, Andre Iguodala, J.J. Barea and Vince Carter. 
This shows that using the information from the tensor components allows us to essentially group players based on different aspects of their game simultaneously. 
Finally, the last cluster includes players that are a mix of the other 4 clusters, which makes it harder to profile them. 
Nevertheless, considering also the location of this cluster on the t-SNE projection, i.e., surrounded by the other four clusters, it further strengthens our belief that {\method} is able to capture multi-aspect patterns in the shooting data.

\begin{figure}[ht]
\centering
\includegraphics[scale=0.45]{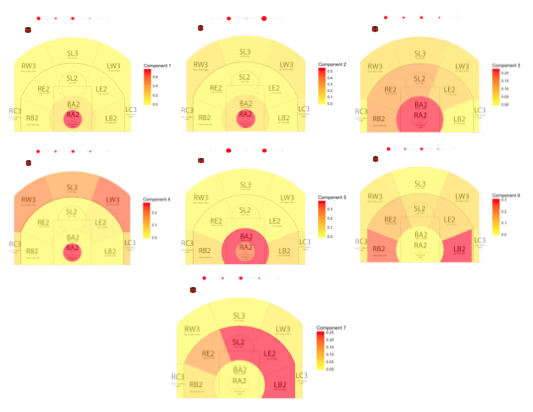}
\caption{{\method} identified 7 components using $\tensor_{Made,Team}$ The teams' offensive tendencies can then be thought of as a combination of these latent patterns.}
\vspace{-0.1in}
\label{fig:shot-team-componenets}
\end{figure}

We also analyzed the teams shooting patterns by designing the appropriate tensor for {\method}. 
In this case tensor $\tensor_{Made,Team}$ is obtained by using the shots from all the players of the teams. 
For the 2014-15 season {\method} identified 7 components, whose spatio-temporal parts are presented in Figure \ref{fig:shot-team-componenets}. 
These patterns can characterize the behavior of teams as a whole - rather than individual players. 
For example, for the Houston Rockets, their made shots  does not include components 1, 3, 6 and 7, i.e., the corresponding coefficients are almost 0. 
Rockets' successful shot selection only follows the latent patterns described in components 2, 4 and 5. 
Note that these components correspond to three-point shots and shots taken from the paint, i.e., the most efficient shots in basketball. 
This is something we should have expected from an analytically savvy team like the Rockets, and hence, {\method} again matches the known intuition and knowledge for the game.
Beyond the pure quality of the models identified by {\method} it is also important to make sure that the patterns identified are \textit{sensible} and our results seem to clearly indicate that.  
This will provide confidence to results obtained when incorporating additional information that has been ignored before and is expected to provide new insights. 
For example, the shot clock information can be an important factor for shot selection. 
When the shock clock winds down, a player will simply take a shot (in most of the cases) to avoid a shot clock violation. 
This shot can be of very bad quality, but the corresponding component obtained from the tensor factorization will inform us for this (i.e., that the shot was taken as the shot clock was expiring). 
The applications of {\method} are only limited by the amount and type of information available to us. 
In the rest of this section, we use {\method} analyze optical tracking data to obtain information about the offensive schemes of teams.

\textbf{Optical Tracking Data Analysis: }
Shooting data can provide a glimpse into the offensive (and defensive depending on the definition of $\tensor$) tendencies of a team and in general of basketball.  
For example, if one tracks the data over different seasons he might observe that the team factor matrix $\mathbf{A}$ for (made) shot charts (Figure \ref{fig:shot-team-componenets}) have larger coefficients for 1, 2, and 4.  
This is a direct consequence of the analytics movement that has identified three-point shots and shots from the paint as the most efficient ones. 
However, the offensive tendencies of the game can be better captured through the detailed optical tracking data described earlier. 
{\method} can be used in this setting as well to identify \textit{prototype} offensive formations.  
For this case the three modes of $\tensor$ correspond to (i) the court zones, (ii) the shot clock\footnote{We have quantized the shot clock information to bins of 1 second.}, and (iii) an identifier for the possession that this snapshot is obtained from. 
The element $\tensor(i,j,k)$ is equal to the number of offensive players on the court zone i, when the shot clock was j during possession k. 
Simply put, $\tensor(i,j,k)$ can take values from 0 to 5. 

\begin{figure}[ht]
\centering
\includegraphics[scale=0.25]{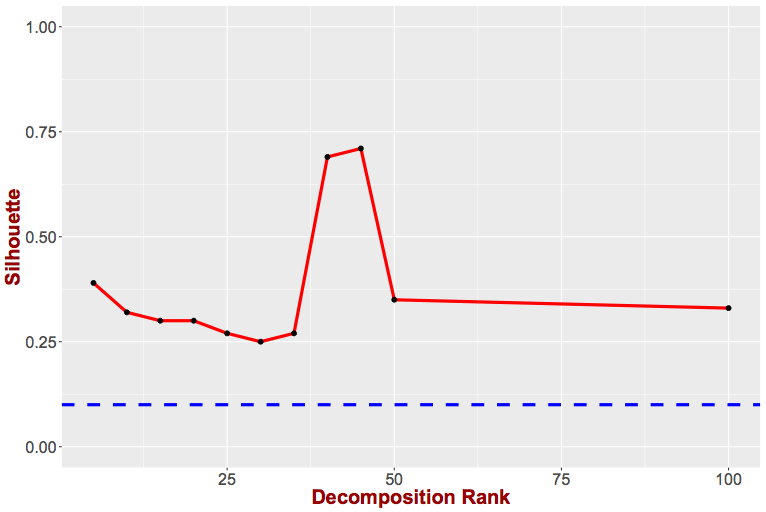}
\vspace{-0.1in}
\caption{While all ranks for the decomposition we examined increase the separability of the possessions as compared to the raw data, a rank of $F=45$ was chosen for this application of {\method} since it exhibits the maximum Shillouette value.}
\label{fig:silhouette}
\vspace{-0.15in}
\end{figure}

We have computed various rank decompositions for $\tensor(i,j,k)$. 
All the possible ranks that CORCONDIA can examine (i.e., up to $F = 13$, which is the minimum dimension of $\tensor$ that corresponds to the court zones) exhibit a good quality model.  
Therefore, we will also utilize the heuristic described in Section \ref{sec:thoops} to choose the decomposition rank.  
In particular, we will cluster the possessions in our dataset. 
Using $K=10$, Figure \ref{fig:silhouette} presents our results, where we have also presented the Silhouette value (horizontal dashed line) for the clustering using the raw data.  
As we see for all the ranks examined the separability obtained from the possession factor matrix is better as compared to that from the raw data. 
The latter exhibit a Silhouette value of approximately 0.1 (horizontal dashed line in Figure \ref{fig:silhouette}), which is typically interpreted as no having identified any substantial structure \cite{kaufman2009finding}. 
When using the decomposition of $\tensor$ for low ranks (e.g., less than 40), the separability, while improved over the raw data, is still fairly low, with Silhoutte values still less than 0.4, which translate to a weak (potentially artificial) structure identified \cite{kaufman2009finding}. 
However, for ranks between 40 and 50 the components are able to identify a reasonably strong structure ($\max{\silhouettes}_{F=40,2:K}=0.69$ and $\max{\silhouettes}_{F=50,2:K}=0.71$). 
Therefore, in this case we can choose $F = 45$. 

\begin{figure*}[ht]
\centering
\includegraphics[scale=0.6]{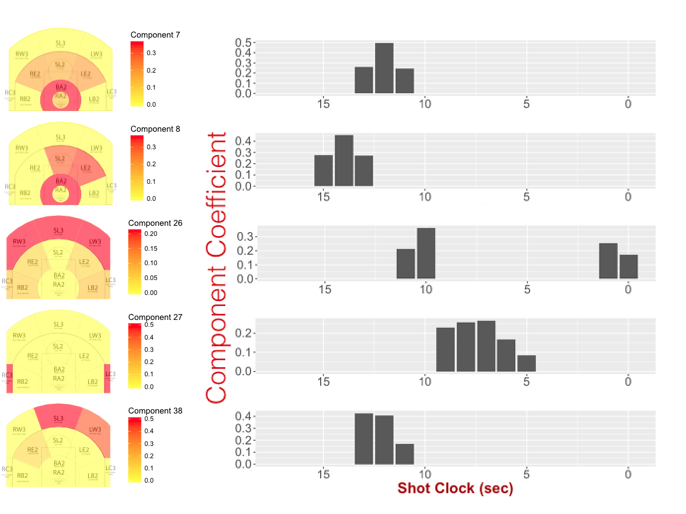}
\vspace{-0.15in}
\caption{Five representative components identified by {\method} from optical data. The spatial part of the component (first column) represents the distribution of the offense's players across the court zones, while the temporal part captures the temporal distribution of this formation over a possession. }
\label{fig:thoops-optical}
\vspace{-0.15in}
\end{figure*}

Due to space limitations we cannot present all 45 components. 
However, Figure \ref{fig:thoops-optical} presents 5 representative components identified for the offensive schemes in the league from {\method}. 
A first observation is that the temporal component (second column) exhibits either a single mode or a bimodal (e.g., fourth row) distribution. 
This is the same for the components omitted and essentially captures the fact that specific formations/schemes are deployed in different stages of the offense.  
The spatial component (first column) provides us with prototype formations for an NBA offense. 
For example, component 27 (fourth row) represents a very common setting the last few years in the NBA, i.e., shooter(s) {\em parked} at the corner 3 area waiting for an assist for the catch-and-shoot attempt. 
Corner 3s have been identified to be one of the most efficient shoots in basketball and hence, team have tried to incorporate this into their offense.

\section{Discussion and Other Applications}
\label{sec:discussion}

In this paper we have presented {\method}, a framework based on tensor decomposition for analyzing multi-aspect basketball data. 
We have showcased its applicability through analyzing spatio-temporal shooting and player tracking data. 
However, the applications of {\method} are not limited only in this type of data. 
Additional information can be integrated into the analysis  through higher order tensors. 
For example, a fourth mode can be included that captures the score differential during the possession. 
This will allow us to identify prototype offensive patterns controlling for the score differential as well. 
Depending on the type of information it might be more appropriate to include them in a matrix coupled with the tensor. 
For instance, personnel information per possession, such as team on offense/defense, player names, boxscore statistics of the players, (adjusted) plus-minus ratings, personal fouls etc., is better represented through a matrix $\mathbf{M}_B$ with rows representing the possessions in the dataset and columns representing different attributes. 
In this case matrix $\mathbf{M}_B$ is coupled with tensor $\tensor$ at the possession dimension and we can obtain the tensor components through a coupled matrix-tensor factorization, which essentially provides a low dimensional embedding of the data in a common contextual subspace by solving a coupled optimization problem. 

However, on-court strategy and scouting is not the only application of {\method}. 
As mentioned earlier, film study is crucial for game preparation but it can be very time consuming to identify the possessions the team wants to study. 
Nevertheless, the components identified by {\method} can drive the development of a system that allows for flexible search in a database of possessions. 
This can automate and facilitate the time-consuming parts related with film study. 
For example, one can imagine querying the system using as input a (probabilistic) spatial distribution for the offense, the shot clock and any other information available for the possessions used to build tensor $\tensor$. 
For instance, let us assume that we want to extract all the possessions where an offense had players positioned at the corner 3 areas during the last 5 seconds of the shot clock for full possessions, i.e., while the shot clock was between 1-5 seconds. 
If one were to use the raw data directly, he would need to go over all the possessions, extract the last 5 seconds and perform spatial queries to examine whether the spatial constraints of the query are satisfied. 
The time complexity for this search is $O(N)$, where $N$ is the number of possessions (i.e., size of data). 
However, using the components identified from {\method} as indices we can further improve the time complexity of this retrieval task, to sub-linear to the size of the data. 
In particular, we can first compute the similarity $\sigma(\mathbf{q},f_i)$, between the query $\mathbf{q}$ - expressed as a vector over the spatial and temporal dimension - and the different components $f_i$ identified from {\method}. 
If $\sigma(\mathbf{q},f_i) > \theta$, for some threshold $\theta$, the system will return the possessions that are mainly represented in component $f_i$, i.e., the possessions with ``high'' coefficient at vector $\mathbf{a}_{f_i}$. 
It should be evident that this process has a time complexity $O(F)$, where $F$ is the number of components. 
Given that the number of components increases slower than the size of the dataset\footnote{An upper bound for the rank of the tensor is $0.5\cdot \min$ {\tt dim}$(\tensor)$, where {\tt dim}$(\tensor)$ is the set of $\tensor$'s dimensions size \cite{sidiropoulos2017tensor}. Given also that we are not interested in the full-rank decomposition of the tensor, the number of $\tensor$'s components used in {\method} will be (much) smaller as well.}, using {\method} for indexing the possession database will significantly accelerate the retrieval of relevant video frames for film study. 
We will further develop the system as part of our future work, where we will also study the various trade-offs between the response time versus precision and recall of the retrieved possessions.

{\method} can also be used to generate synthetic data that exhibit the same patterns as the original data. 
In particular, we can use the components in {\method} as a seed to a dataset generation process, where the coefficients of the component can be normalized to represent a probability distribution function. 
By sampling this probability distributions through Monte Carlo and assigning an actual location for a player within a court zone using a uniform spatial distribution (or a distribution learned from the data), we can generate a synthetic dataset. 
This can be crucial for the research community since many times spatio-temporal data similar to the ones obtained from NBA's optical tracking system are not publicly available. 
{\method} can help bridge the gap between public, open research and private data by allowing the generation of synthetic datasets that exhibit similar patterns with the original data. 
Obviously the synthetic data exhibit similar patterns with the original data only with respect to the dimensions considered in the creation of $\tensor$. 
We will further explore this application as part of our future work as well.

\bibliographystyle{siamplain}
\bibliography{references}

\end{document}

%% file: ex_shared.tex
\usepackage{lipsum}
\usepackage{amsfonts}
\usepackage{graphicx}
\usepackage{epstopdf}
\usepackage{algorithmic}
\ifpdf
  \DeclareGraphicsExtensions{.eps,.pdf,.png,.jpg}
\else
  \DeclareGraphicsExtensions{.eps}
\fi

\usepackage{enumitem}
\setlist[enumerate]{leftmargin=.5in}
\setlist[itemize]{leftmargin=.5in}

\newsiamremark{remark}{Remark}
\newsiamremark{hypothesis}{Hypothesis}
\crefname{hypothesis}{Hypothesis}{Hypotheses}
\newsiamthm{claim}{Claim}

\headers{\method: A Multi-Aspect Analytical Framework for Spatio-Temporal Basketball Data}{Evangelos Papalexakis, Konstantinos Pelechrinis}

\title{\method: A Multi-Aspect Analytical Framework for \\ Spatio-Temporal Basketball Data}

\author{Evangelos Papalexakis\thanks{Department of Computer Science, University of California, Riverside 
  (\email{epapalex@cs.ucr.edu})} \and Konstantinos Pelechrinis\thanks{School of Computing and Information, University of Pittsburgh 
  (\email{kpele@pitt.edu})}}

\usepackage{amsopn}

\makeatletter
\newcommand*{\addFileDependency}[1]{
  \typeout{(#1)}
  \@addtofilelist{#1}
  \IfFileExists{#1}{}{\typeout{No file #1.}}
}
\makeatother

\newcommand*{\myexternaldocument}[1]{%
    \externaldocument{#1}%
    \addFileDependency{#1.tex}%
    \addFileDependency{#1.aux}%
}
